\documentclass[runningheads]{llncs}
\usepackage{graphicx}

\usepackage{booktabs}
\usepackage{multirow}
\usepackage{tikz}
\usepackage{comment}
\usepackage{amsmath,amssymb} 
\usepackage{color}
\newcommand{\parsection}[1]{\textbf{#1} }
\usepackage{cite}
\usepackage{xspace}

\usepackage{makecell}

\usepackage[accsupp]{axessibility}  

\usepackage[pagebackref=true,breaklinks=true,letterpaper=true,citecolor=blue,colorlinks,linkcolor=red,bookmarks=false]{hyperref}

\begin{document}
\pagestyle{headings}
\mainmatter
\def\ECCVSubNumber{939}  

\title{Video Mask Transfiner for High-Quality Video Instance Segmentation}

\titlerunning{Video Mask Transfiner}
%

\author{
 Lei Ke$^{1,2}$ \and Henghui Ding$^1$ \and Martin Danelljan$^1$ \and Yu-Wing Tai$^3$ \and \\ Chi-Keung Tang$^2$ \and Fisher Yu$^1$\\
 }
\authorrunning{Ke et al.}
%
\institute{Computer Vision Lab, ETH Z{\"u}rich \and
The Hong Kong University of Science and Technology \and
Kuaishou Technology\\
\url{http://vis.xyz/pub/vmt}}

\maketitle

\begin{abstract}
While Video Instance Segmentation (VIS) has seen rapid progress, current approaches struggle to predict high-quality masks with accurate boundary details. Moreover, the predicted segmentations often fluctuate over time, suggesting that temporal consistency cues are neglected or not fully utilized. In this paper, we set out to tackle these issues, with the aim of achieving highly detailed and more temporally stable mask predictions for VIS. We first propose the Video Mask Transfiner (VMT) method, capable of leveraging fine-grained high-resolution features thanks to a highly efficient video transformer structure. Our VMT detects and groups sparse error-prone spatio-temporal regions of each tracklet in the video segment, which are then refined using both local and instance-level cues. Second, we identify that the coarse boundary annotations of the popular YouTube-VIS dataset constitute a major limiting factor. Based on our VMT architecture, we therefore design an automated annotation refinement approach by iterative training and self-correction. To benchmark high-quality mask predictions for VIS, we introduce the HQ-YTVIS dataset, consisting of a manually re-annotated test set and our automatically refined training data. We compare VMT with the most recent state-of-the-art methods on the HQ-YTVIS, as well as the Youtube-VIS, OVIS and BDD100K MOTS benchmarks. Experimental results clearly demonstrate the efficacy and effectiveness of our method on segmenting complex and dynamic objects, by capturing precise details.

\keywords{video instance segmentation, multiple object tracking and segmentation, video mask transfiner, iterative training, self-correction}
\end{abstract}

\section{Introduction}







\begin{figure}[!t]
	\centering
	\includegraphics[width=1.0\linewidth]{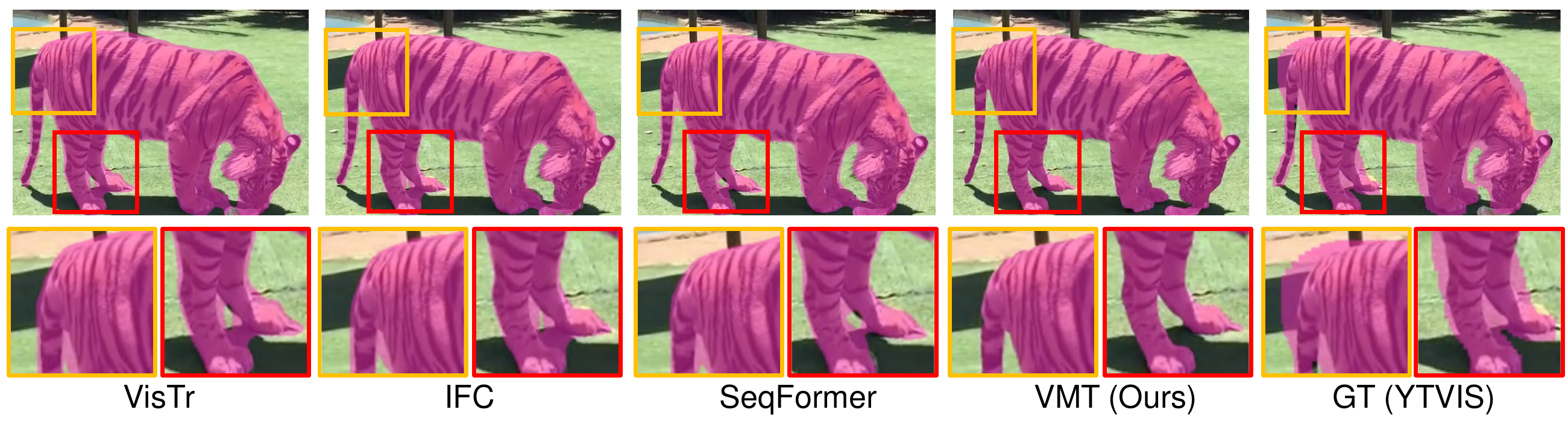}
	\caption{Video instance segmentation results by VisTr~\cite{wang2020end}, IFC~\cite{hwang2021video}, SeqFormer~\cite{wu2021seqformer}, and VMT (Ours) along with the YTVIS Ground Truth. All methods adopt R101 as backbone. VMT achieves highly accurate boundary details, e.g.\ at the feet and tail regions of the tiger, even exceeding the quality of the GT annotations.}
	\label{fig:teaser}
\end{figure}

%
Video Instance Segmentation (VIS) requires tracking and segmenting all objects from a given set of categories. Most recent state-of-the-art methods~\cite{wang2020end,hwang2021video,Fang_2021_ICCV,wu2021seqformer} are transformer-based, using learnable object queries to represent each tracklet in order to predict instance masks for each object. While achieving promising results, their predicted masks suffer from oversmoothed object boundaries and temporal incoherence, leading to inaccurate mask predictions, as shown in Figure~\ref{fig:teaser}. This motivates us to tackle the problem of \emph{high-quality} video instance segmentation, with the aim to achieve accurate boundary details and temporally stable mask predictions. 


Although high-resolution instance segmentation~\cite{kirillov2020pointrend,transfiner} has been explored in the image domain, video opens the opportunity to leverage rich temporal information. Multiple temporal views can help to accurately identify object boundaries, and allow the use of correspondences across frames to achieve temporally consistent and robust segmentation. However, high-quality VIS poses major challenges, most importantly: 1) utilizing long-range spatio-temporal cues in the presence of dynamic and fast-moving objects; 2) the large computational and memory costs brought by high-resolution video features for capturing low-level details; 3) how to fuse fine-grained local features and with global instance-aware context for accurate boundary prediction; 4) the inaccurate boundary annotation of existing large-scale datasets~\cite{yang2019video}. In this work, we set out to address all these challenges, in order to achieve VIS with highly accurate mask boundaries.


\begin{figure}[!t]
	\centering
	\includegraphics[width=0.9\linewidth]{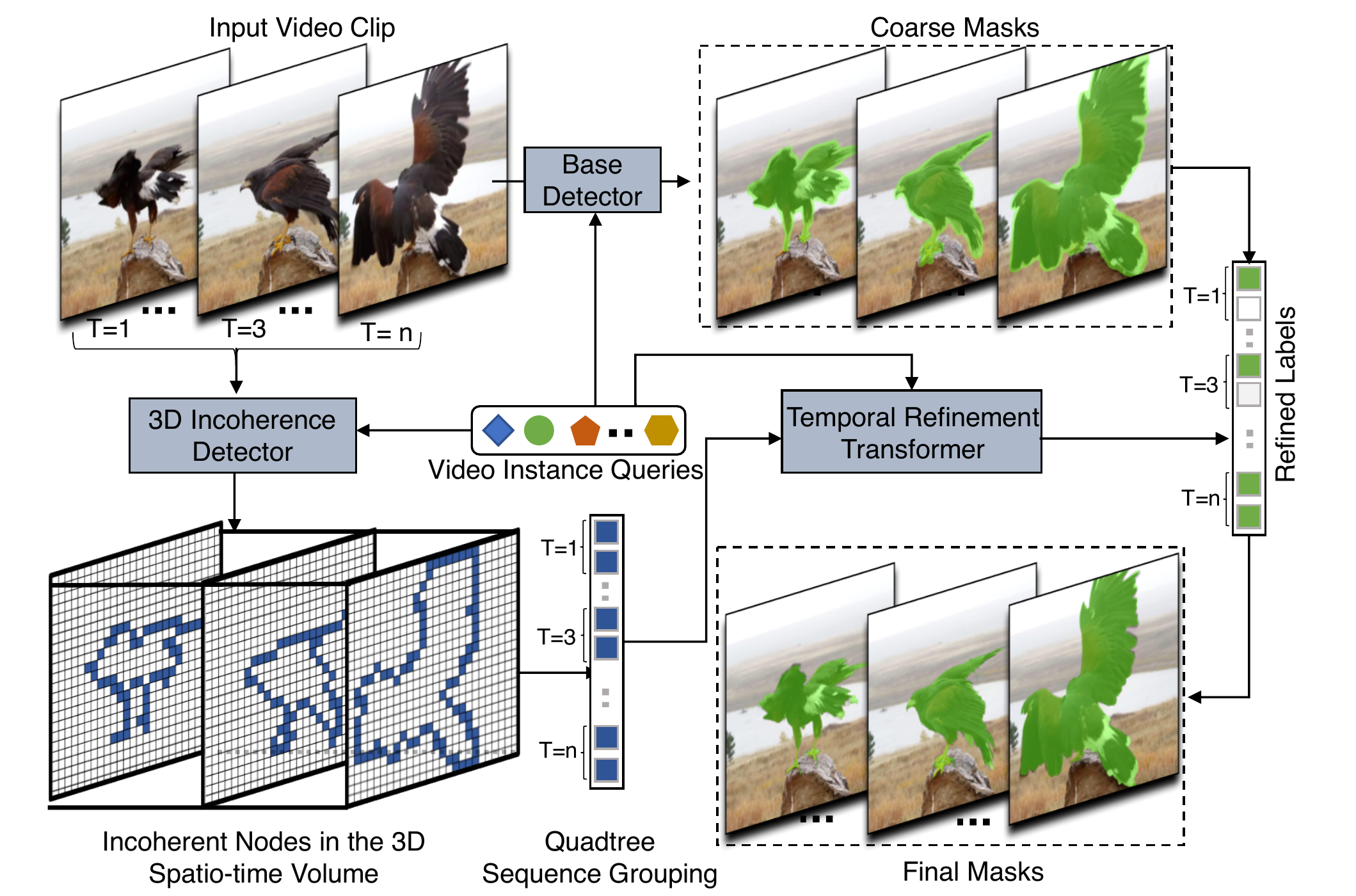}
	\caption{We propose VMT for high-quality video instance segmentation. It adopts a temporal refinement transformer to jointly correct the 3D error-prone regions in the spatio-temporal volume. We employ VMT for automatically correcting the YTVIS with an iterative training paradigm by taking its annotation as coarse masks input. 
	}
	\label{fig:fig2}
\end{figure}

We propose Video Mask Transfiner (VMT), an efficient video transformer that performs spatio-temporal segmentation refinement for high-quality VIS. To achieve efficiency, we take inspiration from Ke~et~al.~\cite{transfiner} and identify a set of sparse error-prone regions. However, as illustrated in Figure~\ref{fig:fig2}, we detect 3D spatio-temporal points, which are often located along object motion boundaries. These regions are represented as a sequence of quadtree points to encapsulate various spatial and temporal scales.
To effectively utilize long-range temporal ques, we group all points and jointly process them using a spatio-temporal refinement transformer. 
Thus, the input sequence for the transformer contains both detailed spatial and temporal information. 
To effectively integrate instance-aware global context, besides using the aggregated points as both input queries and keys of the transformer, we design an additional instance guidance layer (IGL). It makes our transformer aware of both local boundary details and global semantic context.

\begin{figure}[!t]
	\centering
	\includegraphics[width=1.0\linewidth]{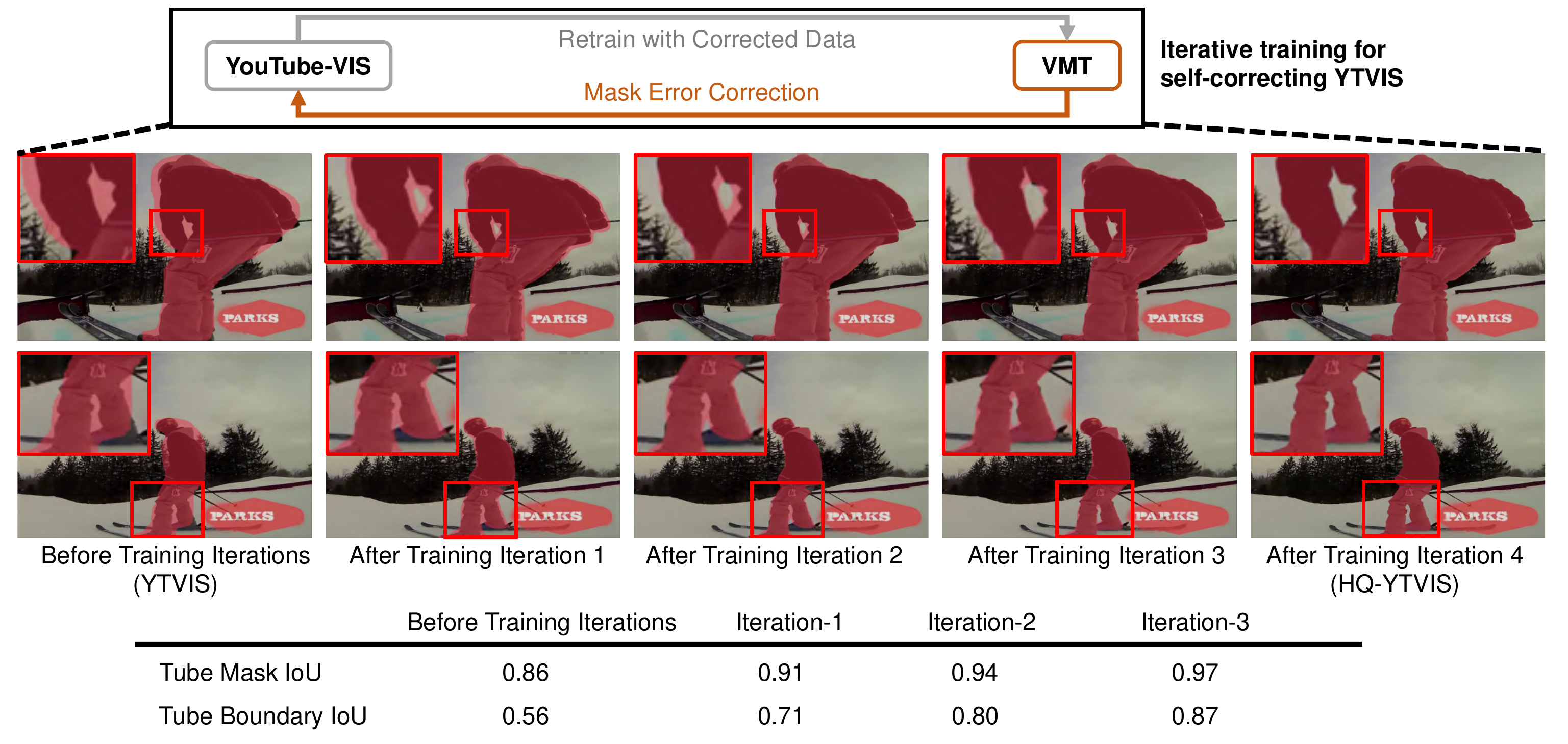}
	\caption{Illustration and intermediate results visualization for iterative training. We show the mask quality change for the given case when correcting YTVIS coarse labels both qualitatively and quantitatively. The predicted instance masks boundaries by VMT becomes more fine-grained with more correction iterations on the YTVIS.}
	\label{fig:loop}
\end{figure}

While our VMT already achieves higher segmentation performance, we observed the boundary quality of the YTVIS~\cite{yang2019video} training annotations to be the next major bottleneck in the strive towards higher-quality mask predictions and evaluation on this popular, large-scale, and highly challenging dataset.
Most importantly, we notice that many videos in YTVIS suffer object boundary inflation issues, as shown in Figure~\ref{fig:teaser} and Figure~\ref{fig:hqytvis}. This introduces a learned bias in the trained model and prohibits very accurate evaluation.
In fact, high-quality training data for VIS is difficult to obtain since dense pixel-wise annotations are costly for a large number videos.
To address this difficulty, instead of manual relabeling the training data, we design an automatic refinement procedure by employing VMT with iterative training. To self-correct mask annotations of YTVIS, both VMT model and training data are alternately evolved, as in Figure~\ref{fig:loop}. To initialize the training of VMT annotation refinement, we use recently proposed OVIS~\cite{qi2021occluded} with better boundary annotations.

To enable benchmarking of high-quality VIS, we introduce the High-Quality YTVIS (HQ-YTVIS) dataset, consisting of our automatically refined training annotations and a manually re-annotated val \& test split.
Moreover, we propose the Tube-Boundary AP evaluation metric that better focuses on segmentation boundary accuracy as well as tracking ability. 
With the proposed HQ-YTVIS dataset, we retrain our VMT and several recent VIS baselines~\cite{yang2019video, wang2020end, pcan, Yang_2021_ICCV,Fang_2021_ICCV,hwang2021video} using our boundary-accurate annotations, providing a comprehensive comparison with current state-of-the-art. We also compare our VMT with state-of-the-art methods on the OVIS~\cite{qi2021occluded} and BDD MOTS~\cite{bdd100k} benchmarks with better annotated boundaries. Quantitative and qualitative results on all three benchmarks demonstrate that VMT not only consistently outperforms existing VIS methods, but also predicts masks at much higher resolution size with small additional computation costs to current video transformer-based methods. We hope our VMT and HQ-YTVIS benchmark could facilitate the community in achieving ever more accurate video instance segmentation.

\section{Related Work}

\parsection{Video Instance Segmentation (VIS)} Extended from image instance segmentation, existing VIS methods can be divided into three categories: two-stage, one-stage, and transformer-based. Earlier methods~\cite{yang2019video,bertasius2020classifying,lin2020video} widely adopted the two-stage Mask R-CNN family~\cite{he2017mask,huang2019mask,ke2021bcnet} by introducing a tracking head for object association. Later works~\cite{CaoSipMask_ECCV_2020,STMask-CVPR2021,liu2021sg} adopted a one-stage instance segmentation framework by using anchor-free detectors~\cite{tian2019fcos} and linear combination of mask bases~\cite{yolact-iccv2019}. For longer temporal information modeling~\cite{lin2021video}, CrossVIS~\cite{Yang_2021_ICCV} proposes instance-to-pixel relation learning and PCAN~\cite{pcan} introduces prototypical cross-attention operations for reading space-time memory. For the transformer-based approach, VisTr~\cite{wang2020end} first uses vision transformer~\cite{carion2020end} for VIS, which is then improved by IFC~\cite{hwang2021video} using memory token communication. Seqformer~\cite{wu2021seqformer} designs query decomposition mechanism. The aforementioned approaches put very limited emphasis on generating very accurate boundary details necessary of high-quality video object masks. In contrast, VMT is the first method targeting for very high-quality video instance segmentation. 

\parsection{Multiple Object Tracking and Segmentation (MOTS)} MOTS methods~\cite{voigtlaender2019mots,milan2015joint,meinhardt2021trackformer} mainly follow the tracking-by-detection paradigm. To utilize temporal features, different from~\cite{Athar_Mahadevan20ECCV,pcan} in clustering/grouping spatio-temporal feature, VMT directly detects the sparse error-prone points in the 3D feature space w/o feature compression and yield highly accurate boundary details. 
	
\parsection{Refinement for Segmentation} Existing works~\cite{kirillov2020pointrend,tang2021look} on instance segmentation refinement are single-image based and thus neglect temporal information. Most of them adopt convolutional networks~\cite{tang2021look} or MLPs~\cite{kirillov2020pointrend}. 
The latest image-based method Mask Transfiner~\cite{transfiner} detects incoherent regions and adopts quadtree transformer for correcting  region errors. Some methods~\cite{takikawa2019gated,cheng2020cascadepsp,ding2019boundary,wang2020deep,yuan2020segfix} focus on refining semantic segmentation details. However, they apply on images without temporal object associations. 

We build VMT based on~\cite{transfiner}, due to its efficiency and accuracy for single image segmentation. The key design of our VMT lies in leveraging temporal information and multi-view object associations of the input video clip. We explore new ways of using video instance queries to detect 3D incoherent points and correct spatio-temporal segmentation errors. Besides, VMT is also a part of our iterative training and self-correction to construct the HQ-YTVIS benchmark. 


\parsection{Self Training}
To reduce the expense of large-scale human-annotation on pixels, some semantic segmentation methods produce pseudo labels for unlabeled data using teacher model~\cite{zhu2020improving,chen2020naive} or data augmentation~\cite{zou2020pseudoseg}. Then, their models are jointly trained on both human-labeled and pseudo labels. In contrast, VMT aims at self-correcting the coarsely or wrongly annotated VIS data. Considering that high-quality VIS requires very accurate video mask annotations to reveal object boundary details, our proposed self-correction and iterative training become even more valuable by eliminating such exhaustive manual labeling.


\section{High-Quality Video Instance Segmentation}

We tackle the problem of high-quality Video Instance Segmentation (VIS), by proposing an efficient temporal refinement transformer, Video Mask Transfiner (VMT), in Section~\ref{sec:VMT}. We further introduce a new iterative training paradigm for automatically correcting  inaccurate annotations of YTVIS in Section~\ref{sec:loop}. 
To facilitate the research in high-quality VIS, we contribute a large-scale HQ-YTVIS benchmark, and propose the Tube-Boundary AP metric in Section~\ref{sec:hqvis}. The proposed benchmark and metric contribute to existing and future VIS models, 
with high-quality annotations for both better training and more precise evaluation.

\subsection{Video Mask Transfiner}
\label{sec:VMT}

\begin{figure}[!t]
	\centering
	\includegraphics[width=0.97\linewidth]{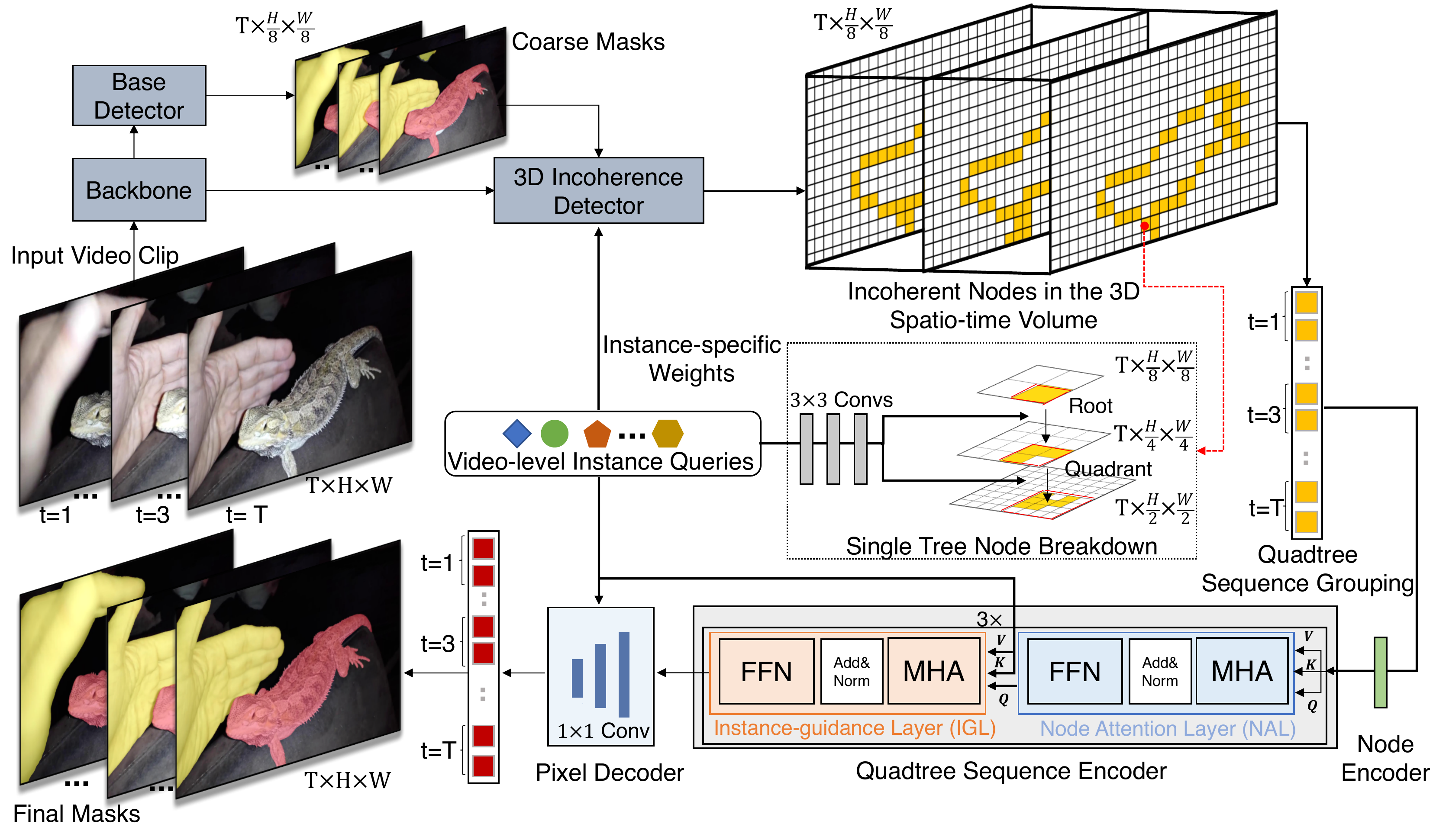}
	\caption{Our VMT framework. A sequence of quadtrees are first constructed in the spatio-time volume by the 3D incoherence detector. Then, these incoherent nodes are concatenated across frames by Quadtree Sequence Grouping. The produced new spatio-temporal node sequences are corrected by temporal refinement transformer under the guidance of video instance queries with global instance context.}
	\label{fig:framework}
\end{figure}

Figure~\ref{fig:framework} depicts
the overall architecture of Video Mask Transfiner (VMT). Our design is inspired by the image-based instance segmentation method Mask Transfiner~\cite{transfiner}. This single-image method first detects  incoherent regions, where segmentation errors  most likely occur in the coarse mask prediction. A quadtree transformer is then used to refine the segmentation in these regions. However, in case of video, the usage of temporal information, including object associations between different frames, is not accounted for by Mask Transfiner. This limits its segmentation performance in the video domain, leading to temporally incoherent mask results. To effectively and efficiently leverage the high-resolution temporal features, we propose three new components for our VMT: 1) an instance query based 3D incoherent points detector; 2) quadtree sequence grouping for temporal information aggregation; and 3) instance query guided incoherent points segmentation. We will describe each of these key components in this section, after a brief summary of the employed base detector in the following.

\subsubsection{Backbone and Base Detector} Given a
video clip that consists of multiple image frames as input, we first use CNN backbone and transformer encoder~\cite{zhu2020deformable} to extract feature maps for each frame. Then, we adopt video-level instance queries to detect and segment objects for each frame following~\cite{wu2021seqformer}. This base detector~\cite{wu2021seqformer} generates initial coarse mask predictions of the video tracklets at low resolution $T\times\frac{H}{8}\times\frac{W}{8}$, where $T$, $H$ and $W$ are the length, height and width of the input video clip. Given this input data, our goal is to predict highly accurate video instance segmentation masks at $T\times H\times W$.

\subsubsection{Query-based 3D Incoherent Points Detection} 
To detect the incoherent regions in the video clip, where segmentation errors are concentrated, a lightweight 3D incoherent region detector is designed. The detector, which encodes the video-level instance query embedding to generate a set of dynamic convolutional weights, consists of  three $3\times3$ convolution layers with ReLU activations. The predicted instance-specific weights are then convolved with the spatio-temporal feature volume at resolution $T\times\frac{H}{8}\times\frac{W}{8}$, followed by a binary classifier to detect the 3D sparse incoherent tree roots. 

We further break down these predicted incoherent points in the 3D volume into each frame. Each point serves as root node in a tree, by branching each node into its four quadrants on the corresponding lower-level frame feature map, which is $2\times$ higher in resolution. The branching is recursive until reaching the largest feature resolution.
We share this 3-layer dynamic instance weights to detect incoherent points for the same video instance across backbone feature sizes at $\{\frac{H}{8}\times\frac{W}{8}, \frac{H}{4}\times\frac{W}{4}, \frac{H}{2}\times\frac{W}{2}\}$, as visualized in Figure~\ref{fig:framework}.
This allows VMT to save a huge computational and memory cost, because only a small part of the high-resolution video features are processed, occupying less than 10\% of the all the points in the 3D temporal volume. 
Video-level instance query captures both positional and appearance information for a time sequence of the same instance in a video clip.
The instance-specific information are already contained in the correlation weights.
Thus, different from~\cite{transfiner}, instance query-based detection removes the necessity of constructing ROI pooling feature pyramid for each video object.
Our 3D incoherent region detector directly operates on the spatio-temporal feature volume from the backbone.

\label{sec:incoherent}

\subsubsection{Quadtree Sequence Grouping}
After detecting 3D incoherent points, we build a sequence of quadtree points within the video clip, each of which resides in a single frame.
To effectively utilize the temporal information across frames, VMT groups together all the tree nodes from all frames of the quadtree sequence, and concatenate them in the token dimension for the transformer. 
The resulting new sequence is the input for the temporal refinement transformer, which contains tree nodes across both spatial and temporal scales, thus encapsulating both detailed spatial and temporal information. We study the influence of different video clip lengths in Table~\ref{tab:length}, which reveals that the input sequence from longer video clips with more diverse and rich information  boosts the accuracy of temporal segmentation refinement.

\subsubsection{Instance Query Guided Temporal Refinement}

For segmenting the newly formed incoherent sequence above, instead of solely leveraging the incoherent points as both input queries and keys~\cite{transfiner}, our Node Attention Layer (NAL) utilizes video-level instance queries as additional semantic guidance. In Figure~\ref{fig:framework}, to inject each point with instance-specific information, we introduce the Instance Guidance Layer (IGL) after each NAL in a level-wise manner. IGL uses incoherent points only as queries, and adopts the video-level instance embedding as the keys and values. This helps our temporal refinement transformer be aware of both
local boundary details and global instance-level context, thus better separating incoherent points among different foreground instances. 
Besides, we add a low-level RGB feature embedding, produced by a network consisting of three 3$\times$3 Conv.\ layers directly operating on the image. This further encapsulates fine-grained object edge details as input to the node encoder. 
Finally, the output is sent into the dynamic pixel decoder for final prediction. 

\subsection{Iterative Training Paradigm for Self-correcting YTVIS}
\label{sec:loop}

\begin{figure}[!t]
	\centering
	\includegraphics[width=1.0\linewidth]{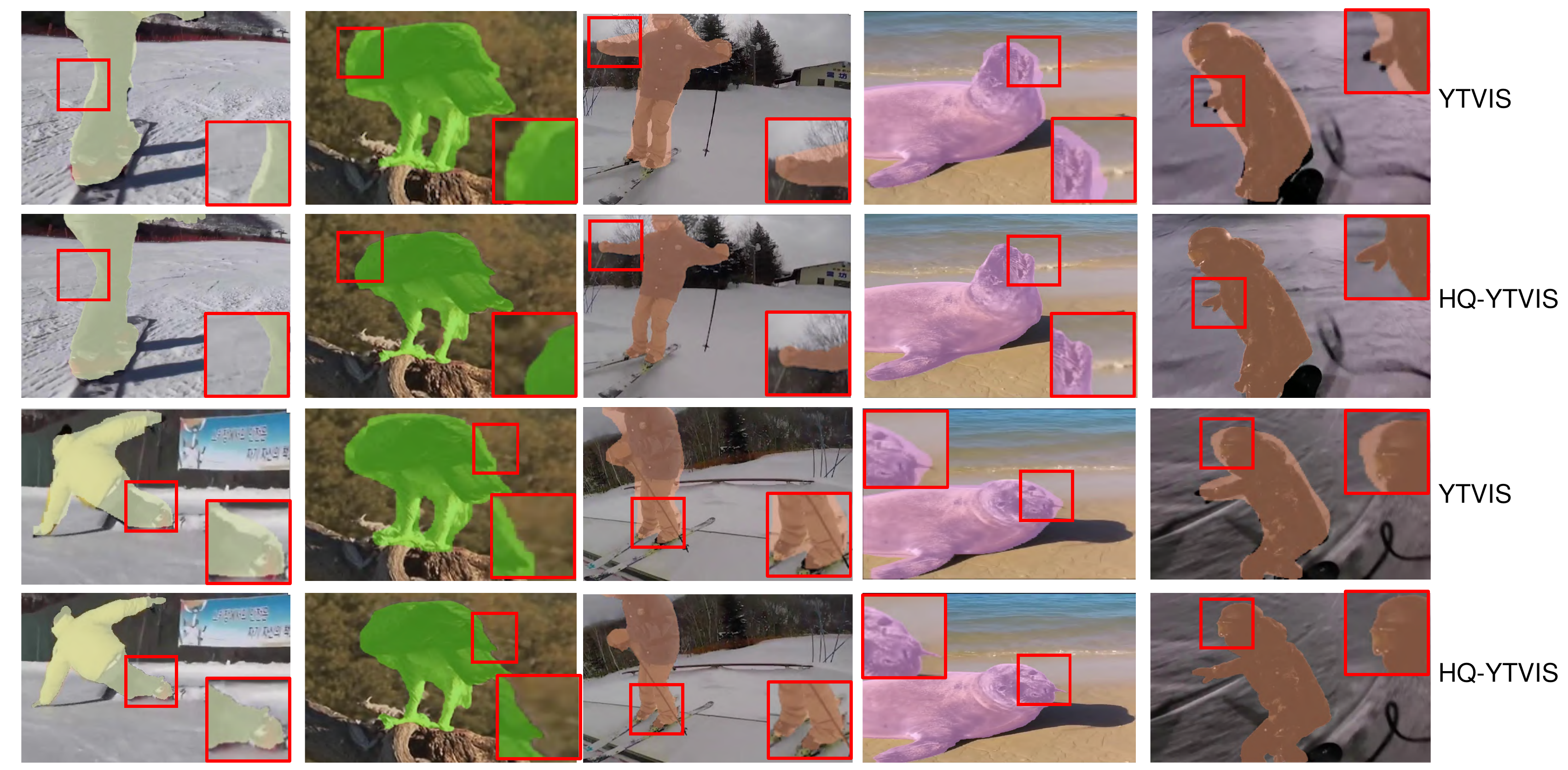}
	\caption{Masks quality comparisons between YTVIS~\cite{yang2019video} and HQ-YTVIS annotations.}
	\label{fig:hqytvis}
\end{figure}

We observed the boundary annotation quality of the YTVIS dataset to be an important bottleneck when aiming to learn highly accurate segmentation masks. 
We show the inaccurate and coarse boundary annotations of YTVIS in Figure~\ref{fig:hqytvis}, Figure~\ref{fig:teaser} and the supplemental video. In particular, we randomly sample 200 videos from the original YTVIS annotations, and find around 28\% of the cases suffer from the boundary inflation problem, where a halo about 5 pixels is around the real object contour. These coarse annotations may due to small number of selected polygon points during instance labeling, which introduces a severe bias in the training, leading to inaccurate boundary prediction.
Based on VMT, we therefore design a method for automatic annotation refinement, and apply it to correct the inaccurate annotations of YTVIS. The core idea is to take the coarse mask annotations from HQ-YTVIS as input and alternate between refining the training data and training the model to achieve gradually improved annotations. 

At the beginning, to equip VMT with initial boundary correction ability, we pretrain VMT on the better annotated OVIS dataset as the first iteration, which has similar data categories and sources as YTVIS. We train the temporal refinement transformer of VMT in a class-agnostic way, leveraging only the incoherent points and video-level instance queries as the input. To simulate various shapes and output of inaccurate segmentation, we degrade the video mask annotations of OVIS~\cite{qi2021occluded} by subsampling the boundary regions followed by random dilations and erosions. Examples of such degraded masks are in the supplemental file. VMT is trained to correct the errors in the ground-truth incoherent regions, and we further enlarge the regions by dilating 3 pixels to introduce both the diversity and the balance of foreground and background pixels ratio in this region.

After training on OVIS, we employ the trained VMT to correct the mask boundary annotations of YTVIS, where the mask annotations of YTVIS are regarded as the coarse mask inputs. We only correct the mask labels when the confidence of the most likely predicted class (foreground or background) is larger than 0.65. Then, we obtain a corrected version of YTVIS and use this new corrected YTVIS data to retrain the temporal refinement transformer of VMT as the 2nd iteration. We iterate this process until the model performance on the manually labeled validation set reaches saturation, requiring 4 iterations. We illustrate the iterative training process and show the intermediate visualizations in Figure~\ref{fig:loop}. After each iteration, the produced annotations masks of YTVIS become more fine-grained until final convergence. We compare the training results using different iterated versions of the YTVIS data, and evaluate their performance on the human-relabeled \textit{val} set in Table~\ref{tab:iters}.

\subsection{The HQ-YTVIS Benchmark}
\label{sec:hqvis}
To facilitate the research in high-quality VIS, we further contribute a new benchmark HQ-YTVIS and design a new evaluation metric Tube-Boundary AP.

\subsubsection{HQ-YTVIS}
To construct the HQ-YTVIS, we first randomly re-split the original YTVIS training set (2238 videos) with coarse mask boundary annotations into train (1678 videos, 75\%), val (280 videos, 12.5\%) and test (280 videos, 12.5\%) subsets following the splitting ratios in YTVIS. 
Then, the masks annotations on the train subset is self-corrected automatically by VMT using iterative training as described in Section~\ref{sec:loop}. The smaller set of validation and test videos are carefully relabeled by human annotators to ensure high mask boundary quality. Figure~\ref{fig:hqytvis} shows the mask annotation differences of the same image from the training set between HQ-YTVIS and YTVIS. HQ-YTVIS has much more accurate object boundary annotations. We retrained VMT and all baselines~\cite{yang2019video, wang2020end, pcan, Yang_2021_ICCV,Fang_2021_ICCV,hwang2021video} on HQ-YTVIS from scratch, and compare the results with those obtained by training them on the original YTVIS annotations with the same set of images. We conduct quantitative results comparisons results in~Table~\ref{tab:data_compare}, which clearly shows the advantage brought by HQ-YTVIS.
We also include the relevant qualitative comparisons in the Supp. file.
We hope HQ-YTVIS can serve a new and more accurate benchmark to facilitate future development of VIS methods aiming at higher mask quality.

\subsubsection{Tube-Boundary AP} 
We propose a new segmentation measure Tube-Boundary AP for high-quality video instance segmentation. The standard tube mask AP in~\cite{yang2019video} is biased towards object interior pixels~\cite{kirillov2020pointrend,cheng2021boundary}, thus falling short of revealing motion boundary errors, especially for large moving objects.
Given a sequence of GT masks $G^{i}_{b...e}$ for instance $i$,  a sequence detected masks $P^{j}_{\hat{b}...\hat{e}}$ for predicted instance $j$, we extend frame index $b$ and $\hat{b}$ to 1, $e$ and $\hat{e}$ to $T$ for temporal length alignment using empty masks. Tube-Boundary AP (AP$^{\text{B}}$) is computed as,
\begin{equation}
    \text{AP}^{\text{B}}(i, j) = \frac{\sum_{t=1}^{t=T} \left | (G^{i}_{t} \cap g^{i}_{t}) \cap (P^{j}_{t} \cap p^{j}_{t}) \right |}{\sum_{t=1}^{t=T} \left | (G^{i}_{t}\cap g^{i}_{t})\cup (P^{j}_{t} \cap p^{j}_{t}) \right |}
\end{equation}
where spatio-temporal boundary regions $g$ and $p$ are respectively the sequential set of all pixels within $d$ pixels distance from the contours of $G^{i}_{b...e}$ and $P^{i}_{\hat{b}...\hat{e}}$ in the video clip.
By definition, Tube-Boundary AP not only focuses on the boundary quality of the objects, but also considers spatio-temporal consistency between the predicted and ground truth object masks. For example, detected object masks with frequent id switches will lead to a low IoU value. 


\section{Experiments}

\subsection{Experimental Setup}

\parsection{HQ-YTVIS \& YTVIS}
We conduct experiments on YTVIS~\cite{yang2019video} and our HQ-YTVIS datasets. YTVIS contains 2,883 videos with 131k annotated object instances belonging to 40 categories. We identify its inaccurate mask boundaries issues in Figure~\ref{fig:hqytvis} and Section~\ref{sec:loop}, which influences both model training and accuracy in testing evaluation.
For HQ-YTVIS, we split the original YTVIS training set (2238 videos) into a new \emph{train} (1678 videos, 75\%), \emph{val} (280 videos 12.5\%) and \emph{test} (280 videos 12.5\%) sets following the ratios in YTVIS. The masks annotations on the \emph{train} subset of HQ-YTVIS is self-corrected by VMT, while the smaller sets of \emph{val} and \emph{test} are carefully relabeled by human annotators to ensure high mask boundary quality. We employ both the standard tube mask \textbf{AP$^M$} in~\cite{yang2019video} and our Tube-Boundary \textbf{AP$^B$} as evaluation metrics. 


\parsection{OVIS} 
We also report results on OVIS~\cite{qi2021occluded}, a recently proposed VIS benchmark on occlusion learning. OVIS has better-annotated boundaries for instance masks with 607, 140 and 154 videos for train, valid and test respectively. 

\parsection{BDD100K MOTS}
We further train and evaluate Video Mask Transfiner on the large-scale BDD100K~\cite{bdd100k} MOTS, which is a self-driving benchmark with high-quality instance masks. It contains 154 videos (30,817 images) for training, 32 videos (6,475 images) for validation, and 37 videos (7,484 images) for testing. 

\subsection{Implementation Details}
Video Mask Transfiner is implemented on the query-based detector~\cite{zhu2020deformable}, and employ~\cite{wu2021seqformer} to provide coarse mask predictions for video instances.
For the temporal refinement transformer, we adopt 3 multi-head attention layers, setting the hidden dimension to 64 and using 4 attention heads. The instance queries are shared between temporal refinement transformer with the base object detector.
During training, we follow the setting in~\cite{wu2021seqformer} and use video clips consisting of 5 frames and sample them from the whole video. We train VMT for 12 epochs and use AdamW~\cite{admw} as optimizer, with initial learning rate set to 2e-4. Our VMT executes at 8.2 FPS on Swin-L backbone. The learning rate is decayed at the 5$^{th}$ and 11$^{th}$ epochs by factor of 0.1. More details are in the Supp. file.

\subsection{Ablation Experiments}

\begin{table}[t]
			\centering%
		\begin{minipage}[t]{0.48\linewidth}
			\centering%
			\caption{Quadtree sequence grouping (QSG) across frames in varying video clip lengths on HQ-YTVIS \textit{val} set.}
			\resizebox{0.83\linewidth}{!}{%
				\begin{tabular}{c | c| c c c | c c c}
					\toprule
					Length & QSG & AP$^B$ & AP$^B_{50}$ & AR$^B_{1}$ & AP$^M$ & AP$^M_{50}$ & AR$^M_{1}$ \\
					\hline
					1 &  & 26.1 & 59.8 & 23.8 & 44.5 & 64.2 & 40.1 \\
				 	\hline
					5 &   & 30.2 & 63.3 & 29.6 & 47.5 & 69.8 & 43.5 \\
					5 & \checkmark & 31.4 & 64.2 & 30.7 & 48.2 & 70.5 & 44.1 \\
				 	\hline
					10 &  & 31.2 & 64.1 & 30.3 & 48.9 & 70.6 & 44.3 \\
					10 & \checkmark & 32.5 & 65.3 & 31.2 & 49.6 & 71.2 & 44.9 \\
				 	\hline
					All &  & 32.3 & 66.0 & 30.6 & 49.7 & 71.8 & 45.5 \\
					All & \checkmark & \textbf{33.7} & \textbf{67.2} & \textbf{31.8} & \textbf{50.5} & \textbf{72.4} & \textbf{46.2} \\
					\bottomrule
				\end{tabular}
			}
			\label{tab:length}
		\end{minipage}
		\label{tab:test1}
		\hfill
		\begin{minipage}[t]{0.48\linewidth}
			\centering%
			\caption{Ablation on 3D incoherent region detector, and refinement region types comparison on HQ-YTVIS validation set. IQ: Instance Query.}
			\resizebox{1.0\linewidth}{!}{%
				\begin{tabular}{c | c | c c c | c c c}
					\toprule
					Region Type & \makecell[c]{Incoherence\\Detector} & AP$^B$ & AP$^B_{50}$ & AR$^B_{1}$ & AP$^M$ & AP$^M_{50}$ & AR$^M_{1}$ \\
					\hline
					\multirow{3}{*}{\makecell[c]{Detected Object \\Boundary}} & FCN & 31.8 & 65.4 & 30.5 & 48.7 & 71.0 & 45.0 \\
					 & IQ (Frame-level) & 31.3 & 64.8 & 29.9 & 48.1 & 69.9 & 44.3 \\
					& IQ (Video-level) & 32.8 & 66.2 & 31.0 & 49.8 & 71.6 & 45.7 \\
					\hline
					\multirow{3}{*}{\makecell[c]{3D Incoherent\\Region}} & FCN & 32.2 & 65.1 & 30.7 & 49.1 & 70.9 & 45.3 \\
					& IQ (Frame-level) & 31.8 & 65.2 & 30.5 & 48.9 & 70.6 & 45.1 \\
				    & IQ (Video-level) & \textbf{33.7} & \textbf{67.2} & \textbf{31.8} & \textbf{50.5} & \textbf{72.4} & \textbf{46.2} \\
				\bottomrule
				\end{tabular}
			}
			\label{tab:3D_inc}
		\end{minipage}
	\end{table}

We conduct detailed ablation studies for VMT using ResNet-101 as backbone on HQ-YTVIS and OVIS \textit{val} sets. We analyze the impact of each proposed component. Besides, we study the effect of iterative training for self-correcting YTVIS, and compare the same models trained on our HQ-YTVIS vs.\ YTVIS.


\parsection{Effect of the Quadtree Sequence Grouping} 
Table~\ref{tab:length} analyzes the influence of video clip lengths to the Quadtree Sequence Grouping (QSG). It reveals that the longer video clips with richer temporal amount indeed brings more performance gain to our VMT. When we increase the tube length from 1 to all frames in the video, a remarkable gain in tube boundary AP$^B$ from 26.1 to 33.7 is achieved. This demonstrate that our approach effectively leverages temporal information, since a tube length 1 performs independent prediction for each frame. Moreover, models \textbf{w/o} QSG are refining the inherent points in each frame separately as~\cite{transfiner}. The multiple boundary view of the same object brings an gain in temporal refinement for over 1.0 AP$^B$.

\parsection{Ablation on the 3D Incoherence Detector}
We study the design choices of our 3D incoherence detector in Table~\ref{tab:3D_inc}. We compare fixed FCN and dynamic FCN (three 3×3 Convs) with weights produced by frame-level or video-level instance queries used in~\cite{wu2021seqformer}. Video-level instance queries achieve the highest AP$^B$, improving 1.9 point compared to the frame-level queries, which shows the effect of temporally aggregated video-level instance information. We also compare 3D incoherent regions with detected object mask boundaries, where the 3D incoherent regions achieves 0.9 AP$^B$ gain. 


\parsection{Effect of Iterative Training} 
In Table~\ref{tab:iters}, we compare MaskTrack~\cite{yang2019video}, SeqFormer~\cite{wu2021seqformer} and VMT for correcting coarse masks of YTVIS in the iterative training. We observe that the improvement scales after each iteration of MaskTrack and SeqFormer on HQ-YTVIS \textit{val} is minor, where the boundary quality AP$^B$ after the 3rd iteration are still coarse (around 60.0 using \textbf{GT} object classes, identities and corresponding coarse masks). In contrast, VMT achieves consistent and large mask quality improvements after three training iterations, which reveals the design advantages of our temporal refinement transformer.



\parsection{Training on YTVIS vs.\ HQ-YTVIS} In Table~\ref{tab:data_compare}, we evaluate the performance of three different approaches when training on either YTVIS or HQ-YTVIS. We train MaskTrack~\cite{yang2019video}, SeqFormer~\cite{wu2021seqformer} and our VMT from scratch with the same set of images.
We use HQ-YTVIS and OVIS for evaluation due to the better annotated mask boundaries. For evaluation on OVIS, we train the mask heads of all these methods in a class-agnostic way, and fix the model weights of the mask head when finetuning them on OVIS for object detection and tracking parts.
All three methods trained using HQ-YTVIS obtain consistent and large performance gain of over 2.0 AP$^B$ on the manually labeled HQ-YTVIS val set, and over 1.0 AP$^M$ on the OVIS val set. 
This shows our self-corrected HQ-YTVIS dataset consistently improves existing VIS methods for segmentation quality, without overfitting to the specific dataset. 

\begin{table}[t]
			\centering%
		\begin{minipage}[t]{0.48\linewidth}
		\centering%
			\caption{Comparison on iterative training. Models after each correction is evaluated on HQ-YTVIS \textit{val} by taking \textbf{GT} classes, ids and coarse masks as input.}
			\resizebox{0.95\linewidth}{!}{%
			\includegraphics[width=1.0\linewidth]{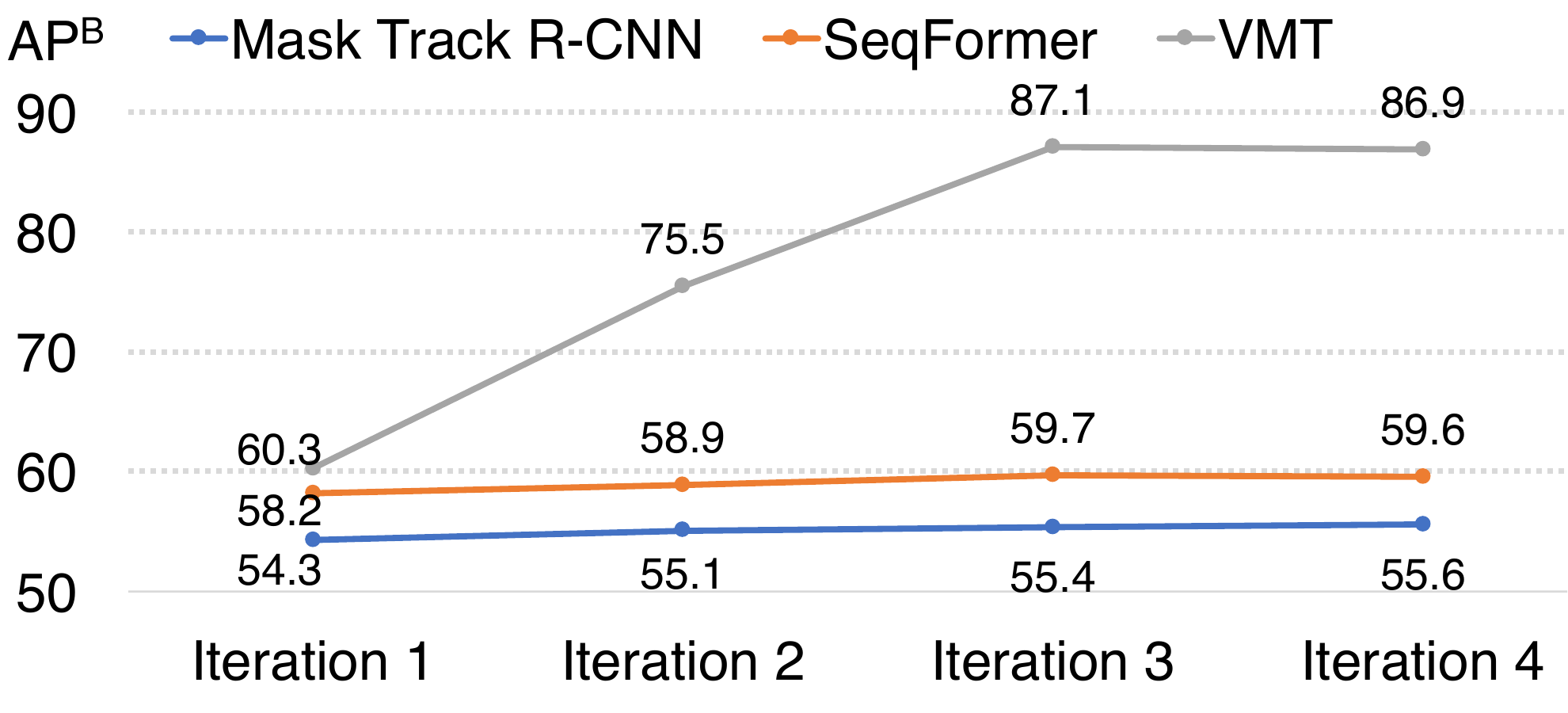}
		}
			\label{tab:iters}
		\end{minipage}
		\label{tab:test1}
		\hfill
		\begin{minipage}[t]{0.5\linewidth}
			\centering%
			\caption{Training on YTVIS \textbf{vs.} HQ-YTVIS with the same images from scratch. We evaluate the trained models  on HQ-YTVIS and OVIS \textit{val} sets.}
			\resizebox{1.0\linewidth}{!}{
		\begin{tabular}{c | c | c | c  c  c | c  c}
		\toprule	\multirow{2}{*}{Method} & \multirow{2}{*}{YTVIS} & \multirow{2}{*}{HQ-YTVIS} & \multicolumn{3}{c|}{HQ-YTVIS} & \multicolumn{2}{c}{OVIS} \\
			
			& & & AP$^B$ & AP$^B_{50}$ & AP$^M$ & AP$^M$ & AP$^M_{50}$ \\
			\midrule
			MaskTrack~\cite{yang2019video} & \checkmark & & 19.8 & 48.9 & 40.2  & 9.3 & 24.2 \\
			MaskTrack~\cite{yang2019video} &  & \checkmark & \textbf{21.7} & \textbf{50.5} & \textbf{41.1}  & \textbf{10.5} & \textbf{25.1} \\
			\midrule
			SeqFormer~\cite{wu2021seqformer} & \checkmark & & 28.9 & 64.2 & 48.6  & 13.8 & 32.1 \\
			SeqFormer~\cite{wu2021seqformer} & & \checkmark & \textbf{31.0} & \textbf{66.1} & \textbf{50.5}  & \textbf{15.2} & \textbf{33.7} \\
			\midrule
			VMT (Ours) & \checkmark & & 30.5 & 64.7 & 48.9 & 15.9 & 33.8 \\
			VMT (Ours) &  & \checkmark & \textbf{33.7} & \textbf{67.2} & \textbf{50.5} & \textbf{17.1} & \textbf{35.0} \\
\bottomrule
		\end{tabular}
	}
			\label{tab:data_compare}
		\end{minipage}
	\end{table}

\begin{figure}[!t]
	\centering
	\includegraphics[width=1.0\linewidth]{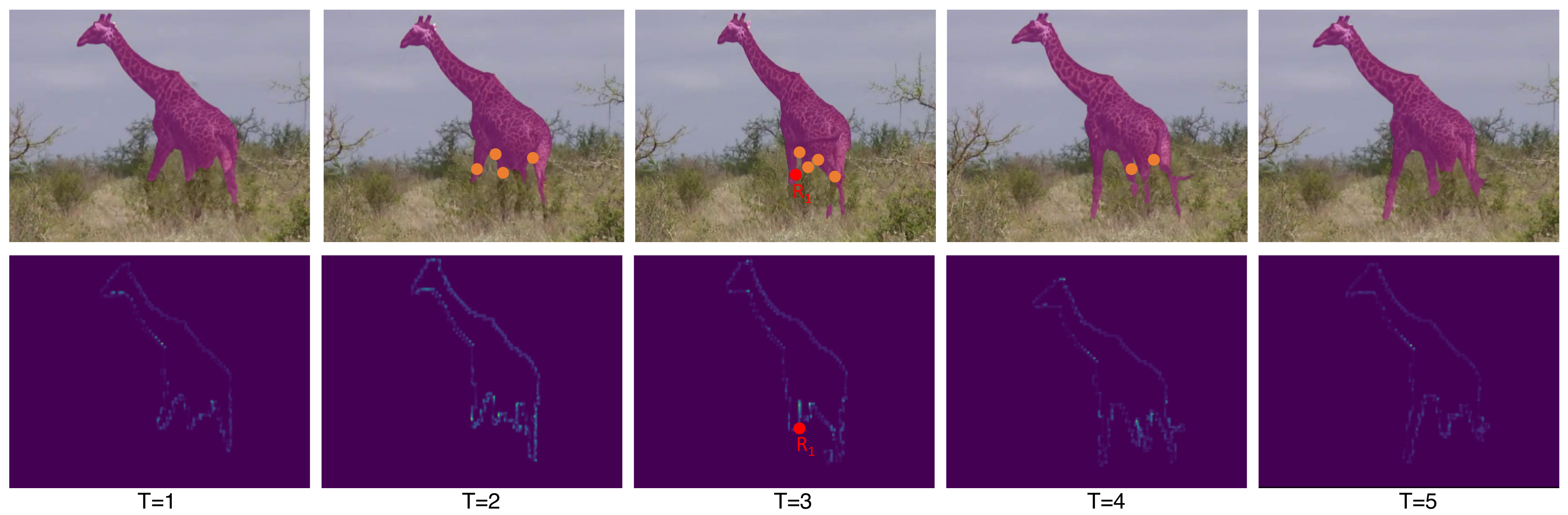}
	\caption{Temporal attention visualizations on the sparse incoherent regions for a video clip of length 5. The sampled red node R1 attends more to the feet regions of the giraffe with semantic correspondence in both the current and neighboring frames. The top 10 attended incoherent node regions are marked in yellow.}
	\label{fig:attention}
\end{figure}

\parsection{Temporal Attention Visualization} 
In Figure~\ref{fig:attention}, we visualize the temporal attention distribution for incoherent nodes in a video-clip of length 5. The attention weights are extracted from the last NAL of the refinement transformer. For the sampled point R1 at T=3, it attends more to the feet regions of the giraffe with semantic correspondence in both the current and neighboring frames. Also, the attention weights for the temporally farther frames are smaller.

\begin{table}[!t]
    \scriptsize
	\centering%
		\caption{ Comparison with state-of-the-art methods on HQ-YTVIS test set and YTVIS~\cite{yang2019video} validation set. All methods, including VMT, are retrained on HQ-YTVIS and YTVIS training sets respectively from scratch for fair comparisons. Results are reported in terms of Tube-Mask AP$^M$~\cite{yang2019video} and our Tube-boundary AP$^B$. VMT predicts mask at output sizes 16$\times$ larger than SeqFormer~\cite{wu2021seqformer}. The advantage of VMT is not fully revealed on YTVIS due to its inaccurate and coarse boundary annotation.}
	\resizebox{0.95\textwidth}{!}{%
	\begin{tabular}{l|c|c|ccc|ccc|ccc}
			\toprule
					\multirow{2}{*}{Method} & \multirow{2}{*}{Backbone} & \multirow{2}{*}{Params} &
					\multicolumn{6}{c|}{HQ-YTVIS} & \multicolumn{3}{c}{YTVIS} \\
					 & & & AP$^B$ & AP$^B_{75}$ & AR$^B_{1}$ & AP$^M$ & AP$^M_{75}$ & AR$^M_{1}$ & AP$^M$ & AP$^M_{75}$ & AR$^M_{1}$ \\
					\midrule
					MaskTrack~\cite{yang2019video} & R50 & 58.1M & 19.8 & 10.6 & 21.1 & 38.8 & 48.6 & 40.3 & 30.3 & 32.6 & 31.0 \\
				    CrossVIS~\cite{Yang_2021_ICCV} & R50 & 37.5M & 23.6 & 16.2 & 24.9 & 43.0 & 52.3 & 44.0 & 36.3 & 38.9 & 35.6 \\
					VisTr~\cite{wang2020end} & R50 & 57.2M  & 24.0 & 16.3 & 25.1 & 43.3 & 52.9 & 44.5 & 36.2 & 36.9 & 37.2 \\
					PCAN~\cite{pcan} & R50 & 36.9M & 23.9 & 16.1 & 25.2 & 42.2 & 51.8 & 43.9 & 36.1 & 39.4 & 36.3 \\ 
					IFC~\cite{hwang2021video} & R50 & 39.3M & 26.5 & 19.6 & 27.5 & 46.6 & 51.5 & 46.9 & 42.8 & 46.8 & 43.8 \\ 
					SeqFormer~\cite{wu2021seqformer} & R50 & 49.3M & 28.6  & 21.4 & 29.3 & 48.5 & 52.2 & 48.5  & 47.4 & 51.8 & 45.5 \\
					\textbf{VMT (Ours)} & R50 & 51.5M & \textbf{30.7} & \textbf{24.2}  & \textbf{31.5} & \textbf{50.5} & \textbf{54.5} & \textbf{50.2} & \textbf{47.9} & \textbf{52.0} & \textbf{45.8} \\
					\midrule
					MaskTrack~\cite{yang2019video} & R101 & 77.2M & 21.7 & 13.1 & 22.8 & 41.1 & 49.1 & 41.7 & 31.8 & 33.6 & 33.2 \\
					CrossVIS~\cite{Yang_2021_ICCV} & R101 & 56.6M & 24.5 & 19.7 & 26.5 & 44.1 & 52.6 & 44.5 & 36.6 & 39.7 & 36.0 \\
					VisTr~\cite{wang2020end} & R101 & 76.3M & 25.1 & 20.5 & 27.7 & 45.3 & 53.2 & 45.1 & 40.1 & 45.0 & 38.3 \\
					PCAN~\cite{pcan} & R101 & 54.8M & 24.8 & 20.1 & 27.0 & 44.0 & 52.1 & 44.3 & 37.6 & 41.3 & 37.2 \\ 
					IFC~\cite{hwang2021video} & R101 & 58.3M  & 27.2 & 23.6 & 28.3 & 48.2 & 51.8 & 47.6 & 44.6 & 49.5 & 44.0 \\ 
					SeqFormer~\cite{wu2021seqformer} & R101 & 68.4M & 30.7 & 27.3 & 30.1 & 49.0 & 52.3 & 46.6  & 49.0 & 55.7 & \textbf{46.8} \\
					\textbf{VMT (Ours)} & R101 & 70.5M & \textbf{32.5} & \textbf{28.9} & \textbf{32.6} & \textbf{51.2} & \textbf{55.1} & \textbf{49.3} & \textbf{49.4} & \textbf{56.4} & 46.7 \\
					\midrule
					SeqFormer~\cite{wu2021seqformer} & Swin-L & 220.0M & 43.3 & 41.5  & 41.6 & 63.7 & 69.7 & 58.7 & 59.3 & 66.4 & 51.7 \\
					\textbf{VMT (Ours)} & Swin-L & 222.3M & \textbf{44.8} & \textbf{43.4}  & \textbf{43.0} & \textbf{64.8} & \textbf{70.1} & \textbf{59.3}  & \textbf{59.7} & \textbf{66.7} & \textbf{52.0} \\
				\bottomrule	
			\end{tabular}}
		\label{tab:viscomp}
	\end{table}
	
\subsection{Comparison with State-of-the-art Methods}
We compare VMT with the state-of-the-art methods on the benchmarks HQ-YTVIS, YTVIS, OVIS and BDD100K MOTS.
Note that we only conduct iterative training when producing the training annotations of HQ-YTVIS. When retraining VMT and all other baselines on the HQ-YTVIS benchmark, all methods are trained from scratch and only once on the same data for fair comparison.

\begin{figure}[!t]
	\centering
	\includegraphics[width=1.0\linewidth]{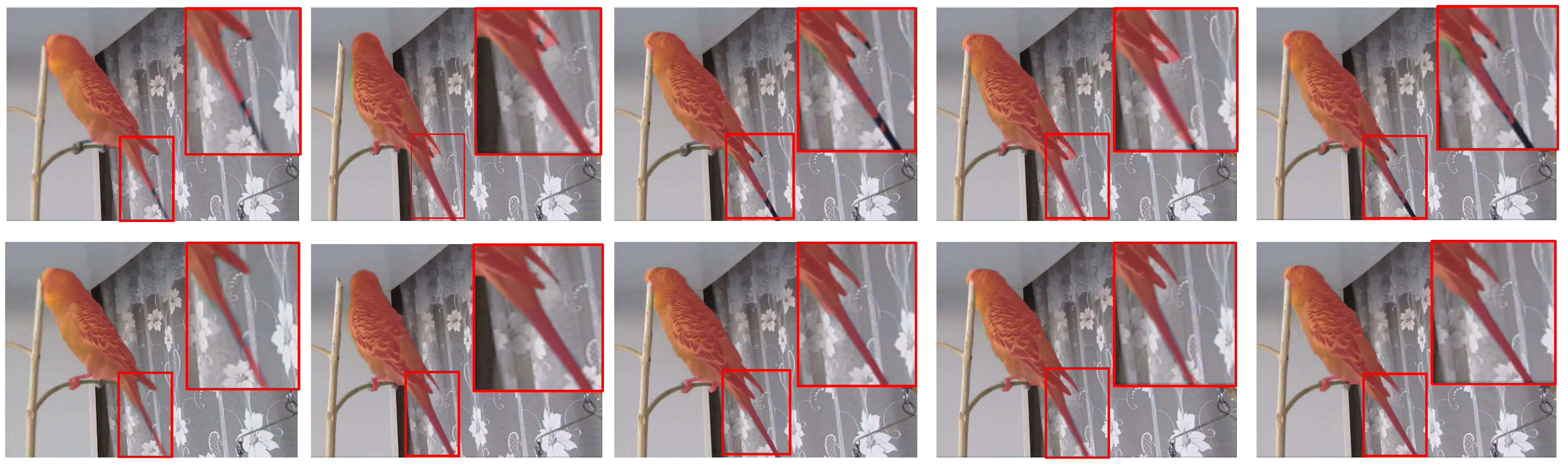}
	\caption{Seqformer (1st row) vs. ours (2rd row) on YTVIS, in terms of mask quality \& temporal consistency. Please refer to the Supp. file for more video results comparisons.}
	\label{fig:vis_comp}
\end{figure}

\parsection{HQ-YTVIS \& YTVIS} Table~\ref{tab:viscomp} compares VMT with state-of-the-art instance segmentation methods on both HQ-YTVIS and YTVIS benchmarks.
VMT achieves consistent performance advantages on different backbones, showing its effectiveness by surpassing SeqFormer~\cite{wu2021seqformer} by around 2.8 AP$^B_{75}$ on HQ-YTVIS using ResNet-50. As in Figure~\ref{fig:hqytvis} and Sec~\ref{sec:loop}, the mask boundary annotation in YTVIS is less accurate. Therefore, the advantages brought by our approach are not fully revealed on this dataset. Yet, VMT exceeds SeqFormer by about 0.5 AP$^M$ on YTVIS with ResNet-50 with higher mask quality as in Figure~\ref{fig:vis_comp}. Moreover, masks predicted by our approach are 16$\times$ larger than those of SeqFormer, while only increasing negligible amount of the model parameters. 

\parsection{OVIS}
The results of OVIS dataset are reported in Table~\ref{tab:ovis}, where VMT achieves the best mask AP 19.8 using Swin-L backbone, improving 1.9 point compared to the baseline SeqFormer~\cite{wu2021seqformer}. 

\parsection{BDD100K MOTS}
Table~\ref{tab:bdd} shows results on BDD100K MOTS, where Mask Transfiner obtains the highest mMOTSA of 28.7 and outperforms the PCAN~\cite{pcan} by 1.3 points by sharing the same object detection tracking heads. The large gain reveals the high quality of temporal masks prediction by VMT. 

	\begin{table}[!t]
			\centering%
		\begin{minipage}[t]{0.42\linewidth}
			\centering%
			\caption{Comparison with state-of-the-art on the OVIS validation set.}
	\resizebox{1.0\textwidth}{!}{%
	\begin{tabular}{lcccccc}
					\toprule
					Method & Backbone & AP & AP$_{50}$ & AP$_{75}$ & AR$_{1}$ & AR$_{10}$ \\
					\hline
					MaskTrack~\cite{yang2019video} & R50  & 10.8 & 25.3 & 8.5 & 7.9 & 14.9 \\
					SipMask~\cite{CaoSipMask_ECCV_2020} & R50 & 10.2 & 24.7 & 7.8 & 7.9 & 15.8 \\
					CrossVIS~\cite{Yang_2021_ICCV} & R50 & 14.9 & 32.7 & 12.1 & 10.3 & 19.8 \\
					STMask~\cite{STMask-CVPR2021} & R50 & 15.4 & 33.8 & 12.5 & 8.9 & 21.3 \\ 
					CMTrack RCNN~\cite{qi2021occluded} & R50 & 15.4 & 33.9  & 13.1 & 9.3 & 20.0 \\ 
					SeqFormer~\cite{wu2021seqformer} & R50 & 15.6 & 34.3 & 12.1 & 9.6 & 21.8 \\
					\textbf{VMT (Ours)} & R50 & \textbf{16.9} & \textbf{36.4}  & \textbf{13.7} & \textbf{10.4} & \textbf{22.7} \\
					\hline
					SeqFormer~\cite{wu2021seqformer} & R101 & 16.2 & 35.1 & 13.1 & 10.3 & 22.9 \\
					\textbf{VMT (Ours)} & R101 & \textbf{17.8} & \textbf{35.7} & \textbf{15.4} & \textbf{10.5} & \textbf{23.8} \\
					\hline
					SeqFormer~\cite{wu2021seqformer} & Swin-L & 17.9 & 35.6 & 15.6 & 10.7 & 24.1 \\
					\textbf{VMT (Ours)} & Swin-L & \textbf{19.8} & \textbf{39.6} & \textbf{17.2} & \textbf{11.2} & \textbf{26.3} \\
					\bottomrule
			\end{tabular}}
			\label{tab:ovis}
		\end{minipage}
		\label{tab:test1}
		\hfill
		\begin{minipage}[t]{0.55\linewidth}
			\centering%
			\caption{State-of-the-art comparison on the BDD100K segmentation tracking validation set using ResNet-50. I: ImageNet. C: COCO. S: Cityscapes. B: BDD100K.}
	\resizebox{1.0\textwidth}{!}{%
				\begin{tabular}{lcccccc}
					\toprule
					Method & Pretrained  & mMOTSA$\uparrow$ & mMOTSP$\uparrow$ & mIDF$\uparrow$ & ID sw.$\downarrow$  & mAP$\uparrow$ \\
					\hline
					SortIoU & I, C, S  & 10.3 & 59.9 & 21.8 & 15951 & 22.2 \\ 
					MaskTrack~\cite{voigtlaender2019feelvos} & I, C, S & 12.3 & 59.9 & 26.2 & 9116 & 22.0 \\
					STEm-Seg~\cite{Athar_Mahadevan20ECCV} & I, C, S  & 12.2 & 58.2 & 25.4 & 8732 & 21.8 \\
					QDTrack~\cite{qdtrack} & I, C, S  & 22.5 & 59.6 & 40.8 & 1340  & 22.4 \\
					QDTrack-fix~\cite{qdtrack} & I, B & 23.5 & 66.3 & 44.5 & 973 & 25.5 \\
					PCAN~\cite{pcan} & I, B  & 27.4 & 66.7 & 45.1 & 876 & 26.6 \\
					\hline
					\textbf{VMT (Ours)} & I, B  & \textbf{28.7} & \textbf{67.3} & \textbf{45.7} & \textbf{825} &  \textbf{28.3} \\
					\bottomrule
			\end{tabular}}
			\label{tab:bdd}
		\end{minipage}
	\end{table}

\section{Conclusion}

We present Video Mask Transfiner, the first high-quality video instance segmentation method. 
Enabled by the efficient video transformer design, VMT utilizes the high-resolution spatio-temporal features for temporal mask refinement and achieves large boundary and mask AP gains on the HQ-YTVIS, OVIS, and BDD100K.
To refine the coarse annotation of YTVIS, we design an iterative training paradigm and adopt VMT to correct the annotations errors of the training data instead of tedious manual relabeling. 
We build the new HQ-YTVIS benchmark with more accurate mask boundary annotations than YTVIS, and introduce Tube Boundary AP for accurate performance measure. 
We believe our method, the new benchmark HQ-YTVIS and evaluation metric will facilitate future video instance segmentation works on improving their mask quality and benefit real-world applications such as video editing~\cite{alldieck2018video,ke2021voin}.

\footnotetext[1]{This work is supported in part by the Research Grant Council of the Hong Kong SAR under grant no. 16201420 and Kuaishou Technology.}



%
%
\bibliographystyle{splncs04}
\bibliography{egbib}
\end{document}